\documentclass{article}

\usepackage{PRIMEarxiv}

\usepackage[utf8]{inputenc}
\usepackage[T1]{fontenc} 
\usepackage{hyperref}
\usepackage{url} 
\usepackage{booktabs}
\usepackage{amsfonts}
\usepackage{nicefrac}
\usepackage{microtype}
\usepackage{lipsum}
\usepackage{fancyhdr}    
\usepackage{graphicx}
\usepackage{subcaption}
\usepackage{amsmath}
\usepackage{amssymb}
\usepackage{booktabs}
\usepackage{amsfonts}
\usepackage{nicefrac}
\usepackage{microtype}
\usepackage{xcolor}
\usepackage{tikz}
\usepackage{tikz-3dplot}
\usepackage{algorithm}
\usepackage{algpseudocode}
\usepackage{amsmath, amsthm, amssymb, amsfonts, mathtools}
\usepackage{multirow}

\usepackage[english]{babel}

\theoremstyle{definition}

\theoremstyle{remark}

\title{Stochastic First-Order Learning for Large-Scale Flexibly Tied Gaussian Mixture Model}

\author{
  Mohammad Pasande, Reshad Hosseini, Babak N. Araabi \\
  School of Electrical and Computer Engineering,\\ University College of Engineering,\\
  University of Tehran, Tehran, Iran\\
  \texttt{\{mohammad.pasande, reshad.hosseini, araabi\}ut.ac.ir} \\
}

\begin{document}
\maketitle

\begin{abstract}
Gaussian Mixture Models (GMMs) are one of the most potent parametric density models used extensively in many applications. Flexibly-tied factorization of the covariance matrices in GMMs is a powerful approach for coping with the challenges of common GMMs when faced with high-dimensional data and complex densities which often demand a large number of Gaussian components. However, the expectation-maximization algorithm for fitting flexibly-tied GMMs still encounters difficulties with streaming and very large dimensional data. To overcome these challenges, this paper suggests the use of first-order stochastic optimization algorithms. Specifically, we propose a new stochastic optimization algorithm on the manifold of orthogonal matrices. Through numerous empirical results on both synthetic and real datasets, we observe that stochastic optimization methods can outperform the expectation-maximization algorithm in terms of attaining better likelihood, needing fewer epochs for convergence, and consuming less time per each epoch.
\end{abstract}

\keywords{GMM, First-Order Optimization, and Manifold Optimization}

\section{Introduction}

\label{sec:intro}
Gaussian Mixture Models (GMMs) are one of the most well-known parametric density models used in a vast set of problems \cite{mclachlan2019finite}.  
Robot movements learning \cite{khansari2011learning}, improving deep learning by expanding batch normalization \cite{kalayeh2019training}, latent variable modeling \cite{kolouri2018sliced}, and even serving as generative models \cite{richardson2018gans} are some of the applications of Gaussian Mixture Models.

In a nutshell, there are two prominent families of approaches to deal with parameter estimation of GMMs, well-known Expectation Maximization (EM) \cite{dempster1977maximum} and numerical optimization \cite{redner1984mixture}. The EM and its extensions \cite{mclachlan2007algorithm} are still more favorable to the community. Each variation has tried to handle some aspects. For instance,  incremental/stochastic adaptions of the EM algorithm have been introduced in  \cite{neal1998view,huda2009stochastic} for online learning, or recently, a hybrid maximization algorithm called Coordinate Descent - Fast Newton Minimum Residual (CD-FNMR) EM has achieved a very fast convergence in many practical examples \cite{asheri2021new}. 
Unlike EM-based methods, most of the numerical optimization approaches have not been fortunate due to two main reasons; firstly, the likelihood maximization can be interpreted as KL-divergence minimization, which does not provide a sufficient sense of distance in far ranges from the optimal point \cite{jin2016local}. Secondly, nonconvex optimization with the implicit Positive Definiteness (PD) constraint on the covariance matrix is challenging. Even though it has been shown that an interior point algorithm can handle a PD constraint through a group of smooth convex inequalities
\cite{vanderbei2000formulating},
with a high dimensional problem, such a method faces slow convergence in comparison with EM-based algorithms.

As a possible remedy of using numerical optimization methods for fitting the parameters of GMMs,  \cite{rahmani2020estimation} came up with a method based on matching the second and third-order moments which is solved by a gradient descent approach. Instead of changing the objective function,  \cite{hosseini2015matrix} addressed the problem of parameter estimation for full covariance matrix by leveraging manifold optimization in a combination of a ``reformulation trick'' to employ sophisticated algorithms such as L-BFGS and nonlinear conjugate gradient. Furthermore,  \cite{hosseini2020alternative} showed that Riemannian Stochastic Gradient Descent (RSGD) outperforms other existing methods. The success key of RSGD in this setup was due to the reformulation trick making the algorithm behave like natural gradient descent \cite{amari1998natural}; the same concept was exploited in  \cite{bouguila2005using} for estimating finite Dirichlet mixture model where the Fisher information matrix is used to alter the gradient descent method. Although recent manifold-powered algorithms can be applied to reasonably high-dimensional GMMs, the full covariance matrices still plea expensive computation especially in the case of a large number of components. Therefore, this paper uses the flexibly-tied GMMs \cite{gales1999semi, asheri2021new} to decrease both the computational costs and the number of components.

Our investigation has three main contributions and motivations:
\begin{itemize}
    \vspace{-0.21cm}
    \item We are first to show that first-order online methods actually work well for fitting flexibly-tied GMMs, outperforming the state-of-the-art batch method of \cite{asheri2021new} in some cases. To avoid singularity in parameter estimation, the common approach of maximizing the penalized likelihood is used.
    \item It was shown in \cite{asheri2021new} that the batch method of \cite{biernacki2006model} for fitting flexibly-tied GMMs with an orthogonality constraint on tied covariances performed inferior to unconstrained structure, and also was very restricted to engage with high-dimensional problems with a large number of components. Thus, a new online first-order optimization algorithm preserving orthogonality constraint is introduced to lift the burden of mentioned tribulations. Interestingly, it is reported that with our proposed algorithm, the constrained structure gives better log-likelihood in some cases.
    \item A favorable side product of using orthogonality constraint is the elimination of determinate computations. However, using the PLU factorization as 
    suggested in \cite{kingma2018glow} can solve the computational costs of computing determinants for unconstrained settings. But, we see that the PLU factorization achieves bad log-likelihood in many cases.
\end{itemize}

From this point forward, in Section \ref{GMM}, we review the flexibly-tied formulation of GMMs; additionally, stochastic first-order gradient-based optimization of nonconvex objective functions is presented in the same section. The manifold of orthogonal matrices is briefly discussed in Section \ref{Manifold}, then a new stochastic orthogonality preserving optimizer is introduced. Experiments on both synthetic and real data, including an empirical investigation of the performance of different methods, are given in Section \ref{exper}. Finally, we conclude our findings and contributions with Section \ref{others}.

\section{Flexibly-Tied Gaussian Mixture Model}
\label{GMM}
The GMM density for \(x\in \mathbb{R}^n\) is simply a convex combination of Gaussian densities, that is
\begin{equation*}
		p(\mathnormal{x}) =\sum^K_{k=1}\pi_k\,\mathcal{N}(\mathnormal{x};\mu_k,\mathnormal{\Sigma}_k^{-1}),
\end{equation*}
where \(\mathcal{N}(.\,; \mu_k, \mathnormal{\Sigma}_k^{-1})\) is a Gaussian density with mean \(\mu_k\) and covariance \(\mathnormal{\Sigma_k}\). 
The common optimization problem for finding the parameters involves minimizing the Negative Log-Likelihood (NLL) of  ``\(l\,\)'' i.i.d. samples, denoted as \(\{x_1, \,x_2,\, \dots, \,x_l\}\), subject to the constraints of the parameters in the density.
\begin{equation}\label{eq:opt_prob}
\begin{aligned}
		& \underset{\pi_k, \mu_k, \Sigma_k}{\text{minimize}}
		& &  - \sum^l_{i=1} \log 	\bigg\{ \sum^K_{k=1}\pi_k\mathcal{N}(\mathnormal{x}_i;\,\mu_k,\mathnormal{\Sigma^{-1}_k)} \bigg\} \\
		& \text{subject to}
		& & \qquad  \sum^K_{k=1}\pi_k = 1,\\
		& & & \quad \mathnormal{\Sigma}^{-1}_{k} \succeq 0\ \& \ \pi_k \geq 0 \qquad\forall \, k.
\end{aligned}
\end{equation} 

The positive definite constraint is a hurdle for using online learning; hence \cite{salakhutdinov2003optimization} suggested using Cholesky factorization for covariance matrices. However, expensive computation and a large number of parameters drive researchers to use simpler forms (like diagonal) for covariance matrices.
The flexibly-tied factorization is suggested in \cite{gales1999semi, asheri2021new} to have the best of both worlds (flexibility and low computation)
\begin{equation}\label{eq:semi-tied}
    \mathnormal{\Sigma}^{-1}_k \;=\; UD_kU^{\mathsf{T}}, 
\end{equation}
where \(U\) is the shared parameter between all components, and \(D_k\) is a component-wise diagonal matrix. 
The PD constraint is restricted to the positiveness of diagonal elements of \(D_k\), which can be satisfied by using a SoftPlus function, that is
\begin{equation*}
    \forall j=\{1,2,\dots,n\}, \quad {d}^{(j)}_k = \frac{1}{\omega}\log \big (  1 + \exp (\omega \tilde{{d}}^{(j)}_k)\big),  \quad 
\end{equation*}
for a constant $\omega$ (in future experiences $\omega = 1$ is used).
Moreover, to handle the constraint of weighting coefficient \(\pi_k\)s, we can change the variables using a SoftMax function (\ref{eq:softmax}) to achieve a new set of variables \(\alpha_k\)s, that is
\cite{jordan1994hierarchical,hosseini2015matrix}
\begin{equation}\label{eq:softmax}
    \alpha_k\,=\,\log\biggl(\frac{\pi_k}{\pi_K}\biggr). 
\end{equation}
\subsection{Regularized Cost Function}
The singularity of the estimates can happen in two cases. In the first case, the probability of one component goes to zero leading to singularity in component weights. The second case is the singularity of the covariance matrix. We regularize the log-likelihood function by using priors for the covariances and component weights (similar to \cite{asheri2021new}). As a result, we have three terms for the regularizer as follows
\begin{itemize}
    \item {
    NLL of the Wishart prior used on \(U\), with the degree of freedom of \(m+2\) and \(S\) as the scale matrix
    \begin{equation}\label{eq:prior_u}
         \psi_1(UU^{\mathsf{T}}) \propto \frac{1}{2}\text{tr}(UU^{\mathsf{T}} S^{-1}) - \frac{1}{2}\text{log}(\text{det}(UU^{\mathsf{T}})),
    \end{equation}
    where \(S\) is proportional to the covariance of the data, that is \(\frac{1}{K^{\frac{2}{n}}}\text{Cov}(X)\). For \(D_k\)s, the NLL of Gamma distribution density is used as the regularizer 
    \begin{equation}\label{eq:prior_d}
        \psi_2(\{D_k\}_{k=1}^K) \propto \sum^K_{k=1}\sum^n_{i=1} \frac{s}{2}d_{ki}-\frac{n}{2}\text{log}(d_{ki}),
    \end{equation}
    where \(s\) is proportional to the sum of diagonal elements of the covariance matrix, that is \(\frac{1}{nK^{\frac{2}{n}}} \sum_{i} \text{Cov}_{ii}(X)\). 
    }
    \item {
    Given the covariance matrices, the conjugate prior for the mean vectors is multivariate Gaussian, therefore we get the following regularizer for the mean vectors
    \begin{equation}\label{eq:prior_mu}
        \psi_3(\{\mu_k\}_{k=1}^K) \propto  \sum^K_{k=1} \frac{\mathcal{K}}{2}(\mu_k-\mu_p)^{\mathsf{T}}{\Sigma}_k^{-1}(\mu_k-\mu_p) +\frac{1}{2}\log\biggl(\text{det}\biggl(\frac{\mathcal{K}}{2\pi}\Sigma^{-1}_k\biggr)\biggr),
    \end{equation}
    where \(\mu_p\) is set to the empirical mean of data, and \(\mathcal{K}\) is the shrinkage parameter set to 0.01.
    }
    \item {
    Symmetric Dirichlet distribution is considered as the prior for the component's weights, and thus, with \eqref{eq:softmax}, we get the following regularizer
    \begin{equation}\label{eq:prior_pi}
        \varphi(\alpha) =    K\zeta \log\bigg(\sum^K_{k=1}\exp(\alpha_k)\bigg)-\zeta\sum^K_{k=1}\alpha_k,
    \end{equation}
     where the \(\zeta\) is the concentration parameter and is set to 0.99. 
    }
\end{itemize}
\subsection{{Stochastic First-Order Learning}}
With all the preparation and considerations, we can proceed to rewrite the minimization problem in \eqref{eq:opt_prob} as an unconstrained optimization problem as below
\begin{equation}\label{eq:new_opt_prob}
\begin{split}
     \underset{\alpha_k, \mu_k, U, D_k}{\text{minimize}} - \sum^l_{i=1} \text{log}	\bigg\{ \sum^K_{k=1}\frac{\text{exp}(\alpha_k)}{\sum^K_{k=1}\text{exp}(\alpha_k)} \mathcal{N}(\mathnormal{x}_i;\,\mu_k,U D_k U^{\mathsf{T}}) \bigg\}\\
     + \psi(U, \, D, \, \mu)\, +\, \varphi(\alpha),
\end{split}
\end{equation} 
where \(\psi(U, \, D, \, \mu)\) is the linear combination of \ref{eq:prior_u}, \ref{eq:prior_d}, and \ref{eq:prior_mu}, that is
\begin{equation}\label{eq:pene}
    \psi(U, \, D, \, \mu) = \psi_1(UU^{\mathsf{T}}) +\psi_2(\{D_k\}_{k=1}^K) + \psi_3(\{\mu_k\}_{k=1}^K).
\end{equation}
One can use First-Order online optimization algorithms like stochastic gradient descent (SGD) with momentum, ADAM \cite{kingma2014adam}, or Clipping SGD \cite{zhang2020adaptive}.   

To follow covariance determinant computation reduction, in \cite{kingma2018glow} the authors introduced a PLU factorization of a matrix where \(L\) is a down triangular matrix with the diagonal element of one, \(\tilde{U}\) is an upper triangular matrix with the diagonal element of zero, and \(s\) is a vector with the motivation of reducing determinant computation. Estimating the \(U\) parameter via this factorization provides the opportunity for reducing the computation cost of the logarithm of determinant in \eqref{eq:semi-tied} as follows
\begin{equation}\label{eq:plu}
U = L( \tilde{U} + \text{diagonal}(s)) \; \Rightarrow \; \text{det}(\Sigma^{-1}_k) = \,\prod_{i}^{} s^2_i\,d_{ki}.
\end{equation}  

Another approach to decrease the computational cost is to enforce orthogonality on matrix \(U\) in \eqref{eq:semi-tied}.  In \cite{celeux1995gaussian}, the authors used EM algorithm to solve flexibly-tied GMM with orthogonality constraint on \(U\). However, \cite{asheri2021new} reported that preserving the orthogonality constraint using the methods developed in \cite{celeux1995gaussian} leads to instability of the optimization algorithm in the case of high-dimensional data and a large number of components.

Enforcing orthogonality constraint on the matrix $U$ in \eqref{eq:semi-tied}
 reduces the determinant computation to the product of the diagonal elements of the matrix \(D_k\). To improve the behavior of optimization algorithm, we use a component-wise multiplicative variable (\(\lambda_k\))  in the covariance factorization, therefore we have
\begin{equation}\label{eq:new_semi-tied}
\Sigma^{-1}_k = \lambda_kUD_kU^{\mathsf{T}} \;\;  \overset{U^{\mathsf{T}}U=I}{\Longrightarrow} \;\; \text{det}(\Sigma^{-1}_k) = \lambda_k^{n}\,\prod_{i}^{} d_{ki}.
\end{equation} 

In contrast to previous works, we suggest using a stochastic manifold-based version of the optimization framework to preserve orthogonality which we introduce in the upcoming section.

\section{Proposed Orthogonality Preserving Optimization}
\label{Manifold}
To solve the objective function (\ref{eq:new_opt_prob}) with the orthogonality constraint on \(U\), this section first reviews the basic prerequisites of Riemannian manifolds. Then, it briefly discusses the geometric properties of orthonormal matrices. Subsequently, we provide a brief overview of the stochastic Riemannian optimization, particularly in the context of handling online data. Finally, we will introduce our novel Riemannian stochastic gradient-based algorithm on the orthonormal matrices, which can be generalized to product manifolds and used for fitting the parameters of flexibly-tied GMMs.
\subsection{Manifold Optimization}

A smooth Manifold ``\(\mathcal{M} \)''  is a space with smooth transitions between its subsets that are locally Euclidean-like. The \textit{Tangent Space} to a smooth Manifold ``\( T_y\mathcal{M} \)'' at a given point (e.g. point ``$y$'') on the manifold is a set containing all the linearized representations of all curves passing through this point; this space is like a vector space approximation to a local neighborhood of $y$. A \textit{Riemannian manifold} ``\( (\mathcal{M}, g) \)''  is a smooth Manifold which is equipped with a \textit{Riemannian metric} ``\( g(\zeta_y, \xi_y) = \langle\zeta_y, \xi_y\rangle_y \)''; a Riemannian manifold is a particular structure allowing the calculus needed to establish optimization algorithms. Riemannian manifold leverages the fact that manifold-aware unconstrained optimization on certain conditions can be seen as an Euclidean constrained optimization problem. Thus, to use this tool to our advantage, several important properties of the manifolds are addressed in the sequel. Details can be found in \cite{absil2009optimization}.

\begin{figure}[h]
\centering
\tdplotsetmaincoords{60}{110}

\pgfmathsetmacro{\radius}{2}
\pgfmathsetmacro{\thetavec}{0}
\pgfmathsetmacro{\phivec}{0}

\begin{tikzpicture}[scale=1.5,tdplot_main_coords, line width=1pt]
\begin{scope}[]
\shade[ball color=blue!20!white,opacity=0.1] (\radius cm,0) arc (0:-180:\radius cm and 0.5*\radius cm) arc (180:0:\radius cm and \radius cm);
\draw[] (0,\radius,0) arc (90:-90:\radius);
\draw[dashed] (0,\radius,0) arc (90:270:\radius);
\end{scope}
\tdplotsetthetaplanecoords{\phivec}
\draw[](0,0,0) 
-- cycle;
\draw[color=red] (0.7*\radius,-1.1*\radius,0.7*\radius) -- (0.5*\radius,-0.1*\radius,1.5*\radius) -- (0.5\radius,0.6*\radius,0.9*\radius) -- (0.5\radius,-0.5*\radius,-0.1*\radius) --
(0.7*\radius,-1.1*\radius,0.7*\radius);
\node[] at (1.4,-1) {\Large \textbf{$\mathcal{M}$}};
\node[red] at (-0.2,-1.5) {\Large \textbf{$T_x\mathcal{M}$}};
\draw[line width=1.2pt, loosely dashed](-0.85*\radius,-0.75*\radius) .. controls (-1.45*\radius,-0.5*\radius) .. (-1.4*\radius,0.2*\radius);

\draw[line width=1pt] (-0.85*\radius,-0.75*\radius) .. controls (-1.45*\radius,-0.5*\radius) .. (-0.8*\radius,0.3*\radius);

\draw[gray,line width=2.9pt, ->] (-0.2*\radius,-0.52*\radius, 0.4*\radius) -- (-0.2*\radius,-0.15*\radius, 0.65*\radius);
\node at (-1.5,-0.7) {\large $\nabla f(y)$};
\draw[gray,line width=1pt, densely dashed] (-0.1*\radius,-0.15*\radius, 1.3*\radius) -- (-0.2*\radius,-0.15*\radius, 0.65*\radius);

\draw[line width=2.5pt, ->] (-0.2*\radius,-0.52*\radius, 0.4*\radius) -- (-0.1*\radius,-0.15*\radius, 1.3*\radius) node[right]{\large $\nabla_E f(y)$};
\fill (canvas cs:x=-12*\radius,y=15*\radius) circle (1.2pt)node[left]{\Large $y$};
\fill (canvas cs:x=16*\radius,y=9*\radius) circle (1.2pt)node[below]{\large $\mathnormal{Exp_y(\nabla f)}$};
\fill (canvas cs:x=19*\radius,y=17.8*\radius) circle (1.2pt)node[right]{\large $\mathcal{R}_y\mathnormal{(\nabla f)}$};

\draw[line width=1.2pt, dashed, blue] (0.2*\radius,-0.2*\radius) .. controls (-0.5*\radius,.5*\radius) .. (0.8*\radius,0.15*\radius) .. controls (0.99*\radius,.22*\radius) .. (0.4*\radius,0.8*\radius)node[midway, below]{\normalsize $\gamma (t)$};
\fill (canvas cs:x=-8.2*\radius,y=-1.8*\radius) circle (1pt)node[blue, left]{\normalsize $t_0$};
\fill (canvas cs:x=17*\radius,y=-9.5*\radius) circle (1pt)node[blue, above]{\normalsize $t_1$};
\draw[line width=1.2pt, blue] (0.2*\radius,-0.2*\radius) .. controls (0.55*\radius,0.5*\radius) .. (0.4*\radius,0.8*\radius)node[midway, above]{\normalsize $\gamma_0 (t)$};

\end{tikzpicture}

\caption{A typical example of a manifold is the sphere, an \(n\)-dimensional manifold embedded in (\({n\!+\!\!1}\!\))-dimensional space, \(\{y \in \mathbb{R}^{n+1} \, | \, \|y\|_2=1 \}\). At any given \(y\), there is a corresponding tangent space (an \(n\)-dimensional vector space which is a hyper-plane approximating the manifold in each point).
The Riemannian metric as \( g : T_y\mathcal{M} \times T_y\mathcal{M} \rightarrow\mathbb{R} \) ) can be easily defined as an inner-product of vectors in this hyper-sphere. The blue dashed line \(\gamma(t)\) is a smooth curve on the manifold between two nominal points, and the solid curve is the so-called geodesic \(\gamma_0(t)\). Note that the \(\nabla_E f\) and \(\nabla f\) indicate the Euclidean gradient and  Riemannian gradient at point \(y\), respectively. The solid black line \(Exp_y(\cdot)\) shows the geodesic curve along the Riemannian gradient vector on the manifold itself, better known as the exponential map. Therefore, the loosely black dashed line \(\mathcal{R}_y\mathnormal{(\cdot)}\) is its compute efficient alternative (retraction function).} \label{fig: manifold}
\end{figure}
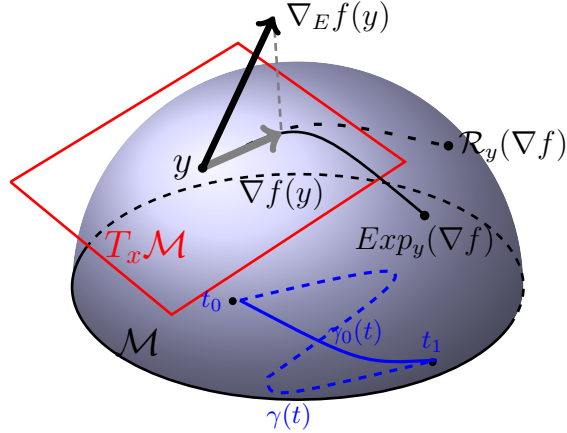

\textit{Geodesic} is the shortest smooth curve connecting two fixed points on a manifold. An \textit{Exponential map} ``\(Exp_y(\cdot)\)'' represents a geodesic along a vector in the tangent space. In most manifolds, the computation of exponential map is expensive, therefore a \textit{Retraction} is preferred ``\(\mathcal{R}_y : T_y\mathcal{M} \rightarrow\mathcal{M} \)''. Retraction is a more general operator that approximates the exponential map (see Figure \ref{fig: manifold}). \textit{Parallel Transport} is used to handle carrying vectors along a geodesic, allowing the ability to move isometrically. However, \textit{Vector Transport} is a smooth mapping on manifold associated with a specific retraction function which replaces parallel transport in practice to gain computation efficiency. ``\(\mathit{w}_y \in  T_y\mathcal{M} \;\text{and}\; \tau \,:\,  T_y\mathcal{M} \times T_x\mathcal{M} \rightarrow T_{\mathcal{R}(\mathit{w}_y)}\mathcal{M}\)''.

\subsection{Orthogonality Constraint}\label{son manifold}
The Stiefel manifold is defined as \(\mathit{p}\) orthogonal columns in $n$-dimensional space, \(\mathit{St}(\mathit{n, p}) = \{ Y \in  \mathbb{R}^{\mathit{n} \times \mathit{p}} \,: \, Y^{\mathsf{T}}Y = I_\mathit{p}\}\). There are two extreme versions of this matrix manifold, first \(\mathit{p}=1\), which creates a sphere in Euclidean space, \(\mathcal{S}^{d-1} = \{ Y \in   \mathcal{E} \, (d =  dim(\mathcal{E})) \,: \, \|y\| = 1\}\); the second version is \(\mathit{p} = \mathit{n}\) that shapes a special orthogonal group: \(\mathcal{O}\mathit{(n)} = \{Y \in  \mathbb{R}^{\mathit{n} \times \mathit{n}} \,: \, Y^{\mathsf{T}}Y = I_\mathit{n}\} \rightarrow \mathcal{SO}\mathit{(n)} = \{X \in \mathcal{O}\mathit{(n)} \,: \, det(Y) = +1\}\).

Some preliminaries are needed to work with the special orthogonal groups; the tangent space is the skew-symmetric matrices which holds the property of vector calculus, and the Riemannian metric is the same as Euclidean metric since we consider \(\mathcal{SO}(n)\) as an embedded submanifold of Euclidean space. Projection onto the tangent space can be obtained by a simple product \(\mathit{Proj_Y}U = AY\); in which the \(A\) represents an auxiliary skew-symmetric matrix as following
\begin{equation}\label{eq:aux_func} 
A \triangleq Skew(UY^{\mathsf{T}}) \quad  or \quad
A \triangleq Skew \Bigl ([I-\frac{1}{2}YY^{\mathsf{T}}]UY^{\mathsf{T}} \Bigl ) \quad
\end{equation}

where the ``\(Skew\)'' function is a typical skew-symmetric form of any matrices multiplied by two, i.e., \(Skew(Y) = (Y - Y^{\mathsf{T}}) \). The choice of retraction function is between various options, but the most famous ones can be summarized as follows: a) \textit{QR} factorization, b) Polar decomposition (using \textit{SVD} decomposition ) \cite{absil2009optimization}, Cayley transform \cite{li2020efficient}. At last, due to the geometry of the problem, a vector transport is just the identity operator.

\subsection{Riemannian Stochastic Optimization}

In many applications, the optimization problem is a finite-sum problem over a Riemannian manifold $\mathcal{M}$, that is
\begin{equation}\label{eq:finit sum} 
\begin{aligned}
& \underset{y \in \mathcal{M}}{\text{min}}	& & f(y)\triangleq \frac{1}{l} \sum_{i=1}^{l} f_i(y).
\end{aligned}
\end{equation}

This finite-sum problem is usually large-scale (in the sense of large number of functions and high dimensionality of $\mathcal M$), therefore the first-order methods suffer from expensive gradient computation. Thus, Riemannian stochastic optimization methods like Riemannian SGD (\cite{bonnabel2013stochastic}) are commonly used in practice.
The Riemannian (minibatch) SGD updates the current estimate $y_{t}$ by
\begin{equation}\label{eq:RSGD} 
y_{t+1} \gets \mathcal{R}_{y_{t}} \left(-\eta_{t}\, {g}_{t}(y_t)\right)
\end{equation}
at each iteration \(t\), where \({g}_{t}\) represents the gradient estimate over the $t$th set of indices (i.e. batch) \(\mathcal B_t\). With proper metric of the manifold, the gradient estimate at each given point can be computed as \({g}_{t}(y_t) = \frac{1}{|\mathcal B_t|} \sum_{j \in \mathcal B_t} \nabla f_{j}(y_t)\). In addition, the \(\eta_t\) is a learning rate (stepsize) typically satisfies \(\sum_{t} \eta_{t} = \infty \;\text{and}\; \sum_{t} \eta_{t}^2 \leq \infty \).

\subsection{Riemannian Stochastic Gradient Clipping}
It is a common assumption to believe the objective function landscape holds an L-smoothness property. 
Therefore we assume the objective function is nonconvex and L-Smooth.
The \textit{Clip} procedure usually includes two significant types Global Clipping ``\textit{GClip}''  and Coordinate-wise Clipping ``\textit{CClip}''; the GClip implies magnitude modifications only with the central assumption of the resemblance of noise in every coordination is the same \cite{zhang2020adaptive}.

 The general stochastic gradient descent with the clipping procedure on \(\mathcal{SO}(n)\) is illustrated in Algorithm \ref{alg:cclip}. The clipping procedure can be written as follows in two versions fixed and adaptive in \(n\)-coordinates in the coordinate-wise sense.

\begin{algorithm}[htbp]
\caption{Stochastic Gradient Clipping on \(\mathcal{SO}(n)\)}\label{alg:cclip}
\begin{algorithmic}[1]

\State $m_t \gets 0$;\Comment{Initialize}

\For{t = 1, \dots , T}
    \State $m_{t+1} \gets \beta_1 m_{t} + (1-\beta_1) {g}_{{E}_t}$ ;  \Comment{\( {g}_{{E}_t}\) : Gradient estimate.}
    \State ${\hat{g}}_{{E}_t} \gets \text{Clip}(\tau_{t+1},\, m_{t+1})$ ; \Comment{Clipping using \ref{eq:clipping}}
    \State ${\hat{g}}_t \gets \mathcal{P}roj({\hat{g}}_{{E}_t})$\Comment{Project onto tangent space using \ref{eq:aux_func}}
    \State $x_{t+1} \gets \mathcal{R}_{x_{t}} (-\eta_{t} \, {\hat{g}}_t)$ \Comment{Retraction, $\eta_{t}$ is the learning rate}
\EndFor
\end{algorithmic}
\end{algorithm}

\begin{equation}\label{eq:clipping} 
\begin{split}
& \textbf{GCLIP}(\tau_{t},\, m_{t}) = \text{min} \{\frac{\tau_{t}}{\|m_{t}\|}, \, 1\}\, m_t, \; \tau_{t} \in \mathbb{R} \geq 0 \quad  \text{or}  \\ 
& \textbf{CCLIP}(\tau_{t},\, m_{t}) = \text{min} \{\frac{\tau_{t}}{|m_{t}|}, \, 1\}\, m_t, \; \tau_{t} \in \mathbb{R}^n \geq 0 \quad  \text{or}  \\ 
& \textbf{ACCLIP}(\tau_{t},\, m_{t}) = \text{min} \{\frac{\tau_{t}}{|m_{t}|+\epsilon}, \, 1\}\, m_t,  \\
& \qquad\qquad\qquad\qquad\qquad \tau_{t}^{\alpha}= \beta_2 \tau_{t-1}^{\alpha} + (1-\beta_2) |{g}_{{E}_t}|^{\alpha}.
\end{split}
\end{equation}

Algorithm \ref{alg:cclip} can be easily generalized to a product manifold of special orthogonal groups ($\mathcal{SO}$)s and Euclidean spaces ($\mathcal{E}$)s. It is enough to replace the ingredients of this algorithm (projection and retraction) with that of the product manifold. The projection and retraction of such a product manifold are equal to the projection and retraction of each element in the product manifold. For the Euclidean space 
 \(\mathcal{E}(n)\) (or \(\mathbb{R}^n\)), its tangent space is also \(\mathcal{E}(n)\) and the projection is identity. Retraction on the Euclidean space is simply the sum of the point and the vector on the tangent space.

The parameters of the flexibly-tied GMM in the optimization problem (\ref{eq:new_opt_prob}) by imposing the new factorization (\ref{eq:new_semi-tied}) form a product of several manifolds (\(\mathcal{SO}(n) \times \prod_{k=1}^{2K} \mathcal{E}(n) \times \mathcal{E}(K)\times \mathcal{E}(K-1)\)), therefore Algorithm \ref{alg:cclip} can be used to solve this problem. The special orthogonal group (\(\mathcal{SO}(.)\)) is used for the matrix \(U\). Also, the Euclidean manifold \(\mathcal{E}(.)\) is employed for the diagonal matrices of  \(D_k\), the component-wise multiplicative variables \(\lambda_k\), the mean vectors  \(\mu_k\), and the Gaussian components' weight  \(\alpha_k\).

\section{Experiments and Results}
\label{exper}
In this Section, we compare several approaches for fitting flexibly-tied GMMs including our proposed method of Section \ref{Manifold} and a Riemannian adaption of ADAM. Also as mentioned in Section \ref{GMM}, there is a gap in comparison of the mentioned methods with gradient-based unconstrained methods; therefore, ADAM and ACClip SGD are considered. Furthermore, the factorization mentioned in \eqref{eq:plu} is exploited in unconstrained methods to evaluate its efficiency to reduce the determinant computation. Since the hybrid algorithm of CD-FNMRS EM has shown to have a fast and trustworthy convergence in a large number of components scheme outperforming EM as reported in \cite{asheri2021new}, the ground truth of our comparison would be CD-FNMRS EM with 27 steps of the coordinate descents algorithm and 73 steps of fast Newton minimum residual method. The total number of iterations in the EM-based algorithm is 100, so the number of epochs of stochastic algorithms is the same. The step size (learning rate) in stochastic algorithms is determined by a cosine annealing procedure with a warm-up\footnote{
The codes for implementations are available via \url{https://github.com/MoPsd/sgd_gmm}.}.

We evaluate the performance of different methods on both synthetic and real datasets. In all tests, 80\% of data were used for training and the rest for testing. All datasets were preprocessed by the whitening procedure, and for real datasets with high dimensions, standard PCA was deployed to reduce the dimensions to not more than 101 with the condition of preserving at least 94\% explained variance. The same initialization of the parameters was used for all methods. 

\begin{table}[t]
\footnotesize
  \centering
    \caption{Obtained errors by different optimization methods for fitting synthetic random data with different number of data-points.}
  \label{tab:sim_rand_data}
  \begin{tabular}{|c |c |c |c |c| c|}
    \hline
    \small{Sep.} & \small{Method} & \small{Size} & \small{Avg. NLL} & \small{Cov. Err} & \small{Mean Err}\\
    \hline
    \footnotesize
    \multirow{15}{*}{\footnotesize{High}} & \multirow{3}{*}{\footnotesize{Adam Euclidean}} & 250 & 2.516 & 8.932 & 0.825\\
    & & 2500 & 0.409 & 5.671 &  0.656 \\
    & & 25000 & -0.247 & 1.019 & 0.417 \\
    \cline{2-6}
	& \multirow{3}{*}{\footnotesize{ACClip Euclidean}} & 250 & 2.486 & 8.268 & 0.676\\
    & & 2500 & -0.029 & 3.770 & 0.352 \\
    & & 25000 & -0.382 & 3.066 & 0.339 \\
    \cline{2-6} 
	& \multirow{3}{*}{\footnotesize{Adam Manifold}} & 250 & 4.581 & 8.320 & 0.773\\
    & & 2500 & 3.233 & 4.427 & 0.108 \\
    & & 25000 & 3.261 & 4.370 & 0.111 \\
    \cline{2-6}  
	& \multirow{3}{*}{\footnotesize{ACClip Manifold (proposed)}} & 250 & 1.691 & 4.874 & 0.444\\
    & & 2500 & -0.024 & 2.140 & 0.201 \\
    & & 25000 & -0.444 & 0.793 & 0.0001 \\
    \cline{2-6} 
	& \multirow{3}{*}{\footnotesize{CD-FNMRS}} & 250 & 0.022 & 2.249 & 0.336\\
    & & 2500 & -0.073 & 2.290 & 0.223 \\
    & & 25000 & -0.404 & 1.808 & 0.099 \\
    \hline    
    \multirow{15}{*}{\footnotesize{Mid}} & \multirow{3}{*}{\footnotesize{Adam Euclidean}} & 250 & 6.948 & 7.88 & 1.448\\
    & & 2500 & 6.336 & 4.895 & 0.609 \\
    & & 25000 & 6.028 & 4.263 & 0.3959 \\
    \cline{2-6}
	& \multirow{3}{*}{\footnotesize{ACClip Euclidean}} & 250 &  6.941 & 7.831 & 1.428\\
    & & 2500 & 6.306 & 4.617 & 0.476 \\
    & & 25000 & 6.050 & 4.557 & 0.477 \\
    \cline{2-6} 
	& \multirow{3}{*}{\footnotesize{Adam Manifold}} & 250 & 6.989 & 7.957 & 1.330\\
    & & 2500 & 6.662 & 6.552 & 0.674 \\
    & & 25000 & 6.592 & 6.101 & 0.574 \\
    \cline{2-6}  
	& \multirow{3}{*}{\footnotesize{ACClip Manifold (proposed)}} & 250 & 6.800 & 7.666 & 1.380\\
    & & 2500 & 6.468 & 5.388 & 0.576 \\
    & & 25000 & 6.245 & 5.160 & 0.422 \\
    \cline{2-6} 
	& \multirow{3}{*}{\footnotesize{CD-FNMRS}} & 250 & 6.698 & 4.880 & 0.662\\
    & & 2500 & 6.266 & 3.642 & 0.275 \\
    & & 25000 & 6.049 & 3.591 & 0.300 \\
    \hline
    \multirow{15}{*}{\footnotesize{Low}} & \multirow{3}{*}{Adam Euclidean} & 250 & 6.991 & 11.34 & 1.831\\
    & & 2500 & 6.980 & 7.579 & 1.717 \\
    & & 25000 & 6.782 & 6.971 & 1.444 \\
    \cline{2-6}
	& \multirow{3}{*}{\footnotesize{\footnotesize{ACClip Euclidean}}} & 250 & 6.991 & 11.37 & 1.831\\
    & & 2500 & 6.983 & 7.633 & 1.734 \\
    & & 25000 & 6.777 & 7.047 & 1.433 \\
    \cline{2-6} 
	& \multirow{3}{*}{\footnotesize{Adam Manifold}} & 250 & 7.006 & 11.45 & 1.830\\
    & & 2500 & 7.063 & 8.660 & 1.705 \\
    & & 25000 & 6.953 & 9.369 & 1.825 \\
    \cline{2-6}  
	& \multirow{3}{*}{\footnotesize{ACClip Manifold (proposed)}} & 250 & 7.049 & 11.42 & 1.831\\
    & & 2500 & 6.983 & 7.967 & 1.685 \\
    & & 25000 & 6.802 & 7.410 & 1.500 \\
    \cline{2-6} 
	& \multirow{3}{*}{\footnotesize{CD-FNMRS}} & 250 & 7.084 & 7.650 & 2.359\\
    & & 2500 & 6.967 & 4.957 & 1.618 \\
    & & 25000 & 6.780 & 5.104 & 1.056 \\
    \hline    
  \end{tabular}
\end{table}
\begin{table}[t]
\footnotesize
  \centering
    \caption{Obtained errors by different optimization methods for fitting  synthetic orthogonal data with different number of data-points.}
  \label{tab:sim_orth_data}
  \begin{tabular}{|c |c |c |c |c| c|}
    \hline
    \small{Sep.} & \small{Method} & \small{Size} & \small{Avg. NLL} & \small{Cov. Err} & \small{Mean Err}\\
    \hline
     \footnotesize
    \multirow{15}{*}{\footnotesize{High}} & \multirow{3}{*}{\footnotesize{Adam Euclidean}} & 250 & 4.472 & 10.49 & 1.206\\
    & & 2500 & 3.222 & 5.951 & 0.822 \\
    & & 25000 & 3.326 & 5.351 & 1.073 \\
    \cline{2-6}
	& \multirow{3}{*}{\footnotesize{ACClip Euclidean}} & 250 & 4.445 & 10.39 & 1.238\\
    & & 2500 & 3.206 & 5.776 & 0.731 \\
    & & 25000 & 3.238 & 5.029 & 0.754 \\
    \cline{2-6} 
	& \multirow{3}{*}{\footnotesize{Adam Manifold}} & 250 & 5.322 & 7.796 & 1.015\\
    & & 2500 & 4.778 &  5.748 & 0.690 \\
    & & 25000 & 4.856 & 6.608 & 0.907 \\
    \cline{2-6}  
	& \multirow{3}{*}{\footnotesize{ACClip Manifold (proposed)}} & 250 & 3.941 &  5.077 & 0.747\\
    & & 2500 & 3.418 & 4.024 & 0.434 \\
    & & 25000 & 3.279 & 4.221 & 0.438 \\
    \cline{2-6} 
	& \multirow{3}{*}{\footnotesize{CD-FNMRS}} & 250 & 3.436 & 4.486 & 0.753\\
    & & 2500 & 3.187 & 4.406 & 0.569\\
    & & 25000 & 3.235 & 5.098 & 0.848 \\
    \hline  
    \multirow{15}{*}{\footnotesize{Mid}} & \multirow{3}{*}{\footnotesize{Adam Euclidean}} & 250 & 7.144 & 8.967 & 2.317\\
    & & 2500 & 7.100 & 8.392 & 1.947 \\
    & & 25000 & 7.085 & 7.517 & 0.933 \\
    \cline{2-6}
	& \multirow{3}{*}{\footnotesize{ACClip Euclidean}} & 250 & 7.145 & 8.927 & 2.330\\
    & & 2500 & 7.099 &  8.383 & 1.928 \\
    & & 25000 & 7.085 & 7.493 & 0.964 \\
    \cline{2-6} 
	& \multirow{3}{*}{\footnotesize{Adam Manifold}} & 250 & 7.153 & 8.915 & 2.244\\
    & & 2500 & 7.095 & 8.722 & 1.999 \\
    & & 25000 & 7.089 & 7.635 & 1.103 \\
    \cline{2-6}  
	& \multirow{3}{*}{\footnotesize{ACClip Manifold (proposed)}} & 250 & 7.186 & 7.89 & 2.167\\
    & & 2500 & 7.100 & 7.933 & 1.794 \\
    & & 25000 & 7.087 & 7.268 & 1.500 \\
    \cline{2-6} 
	& \multirow{3}{*}{\footnotesize{CD-FNMRS}} & 250 & 7.308 & 7.071 & 2.076\\
    & & 2500 & 7.112 & 8.199 & 1.728 \\
    & & 25000 & 7.088 & 8.589 & 1.77 \\
    \hline
\multirow{15}{*}{\footnotesize{Low}} & \multirow{3}{*}{Adam Euclidean} & 250 & 7.118 & 8.365 & 3.345\\
    & & 2500 & 7.111 & 8.392 & 2.358 \\
    & & 25000 & 7.108 & 8.238 & 2.292 \\
    \cline{2-6}
	& \multirow{3}{*}{\footnotesize{\footnotesize{ACClip Euclidean}}} & 250 & 7.123 & 8.339 & 3.339\\
    & & 2500 & 7.110 & 8.388 & 2.359 \\
    & & 25000 & 7.108 & 8.210 & 2.297 \\
    \cline{2-6} 
	& \multirow{3}{*}{\footnotesize{Adam Manifold}} & 250 & 7.097 & 8.516 & 3.257\\
    & & 2500 & 7.108 & 8.711 & 2.353 \\
    & & 25000 & 7.108 & 8.621 & 2.275 \\
    \cline{2-6}  
	& \multirow{3}{*}{\footnotesize{ACClip Manifold (proposed)}} & 250 & 7.102 & 7.908 & 3.207\\
    & & 2500 & 7.117 & 8.257 & 2.303 \\
    & & 25000 & 7.107 & 8.552 & 2.339 \\
    \cline{2-6} 
	& \multirow{3}{*}{\footnotesize{CD-FNMRS}} & 250 & 7.312 & 6.952 & 3.127\\
    & & 2500 & 7.123 & 8.339 & 2.070 \\
    & & 25000 & 7.108 & 8.768 & 2.151 \\
    \hline    
  \end{tabular}
\end{table}
\begin{table}[t]
\footnotesize
  \centering
    \caption{The performance of different methods for fitting real data. The values in the parentheses are the dimensionalities of data.}
  \label{tab:peformance real data}
  \begin{tabular}{|c|c|c|c|}
    \hline
    \small{  Dataset (n)} & \small{Method} & \small{NLL} & \small{Time per Epoch} \\
    \hline
    \footnotesize
    \multirow{6}{*}{MAGIC (10)} & Adam Euclidean & 9.56 & $0.88\pm0.02s$  \\ 
	& ACClipEuclidean& \textbf{9.54}  & $1.05\pm 0.03s$  \\ 
	& Adam Euclidean PLU & 9.64   & $0.89 \pm 0.03s$ \\  
	& Adam Manifold$^{\#}$& 10.85  & $0.85\pm 0.03s$ \\  
	& ACClip Manifold (proposed) & 9.93 & $1.03\pm0.03s$ \\ 
	& CD-FNMRS & 9.62 & \boldmath{$0.15\pm0.03s $} \\ \hline
    \footnotesize
    \multirow{6}{*}{WAVE (21)} & Adam Euclidean & 29.57 & $0.49\pm0.01s$  \\ 
	& ACClip Euclidean & 29.58  & $0.57\pm 0.01s$  \\ 
	& Adam Euclidean PLU & 29.52   & $0.48 \pm 0.01s$ \\  
	& Adam Manifold& 29.99  & $0.43\pm 0.01s$ \\  
	& ACClip Manifold (proposed) & \textbf{29.51} & $0.65\pm0.02s$ \\ 
	& CD-FNMRS & 29.53 & \boldmath{$0.23\pm0.03s $} \\ \hline
    \multirow{6}{*}{USPS (65)} & Adam Euclidean & 62.51 & \boldmath{$4.43\pm0.44s$} \\
	& ACClip Euclidean & \textbf{59.63}  & $4.75 \pm  0.37s$ \\
	& Adam Euclidean PLU & 70.22 & $4.64 \pm 0.72s$ \\
	& Adam Manifold & 78.47  & $3.77\pm 0.40s$ \\  
	& ACClip Manifold (proposed) & 62.52 & $4.62 \pm0.37s$\\  
	& CD-FNMRS & 59.66  & $15.75 \pm8.37s$\\ \hline
    \footnotesize
    \multirow{6}{*}{YEAR (90)} & Adam Euclidean & 98.16 & $94.63 \pm5.53s$  \\
	& ACClip Euclidean & \textbf{97.31}  & $103.34 \pm 5.87s$  \\
	& Adam Euclidean PLU & 100.30   & \boldmath{$83.50 \pm 7.96s$}  \\
	& Adam Manifold & 101.61  & $78.13\pm  4.07s$ \\
	& ACClip Manifold (proposed) & 98.22 & $97.37 \pm3.72s$ \\
	& CD-FNMRS & 97.74  & $141.06 \pm 60.20s$ \\ \hline  
    \multirow{6}{*}{
     \footnotesize
    SVHN (100)} & Adam Euclidean & 83.90 & $83.37 \pm6.49s$ \\ 
	& ACClip Euclidean & 85.95  & $90.41 \pm 6.54s$ \\ 
	& Adam Euclidean PLU & 82.40   & \boldmath{$79.70 \pm 11.47s$} \\
	& Adam Manifold & 74.64  & $86.04 \pm  3.19s$ \\
	& ACClip Manifold (proposed) & \textbf{63.57} & $85.01 \pm 6.27s$ \\
	& CD-FNMRS & 72.17  & $216.30 \pm 107.42s$ \\   
    \hline
    \multirow{6}{*}{
     \footnotesize
    STL (101)} & Adam Euclidean & 99.04 & $106.47 \pm6.02s$ \\ 
	& ACClip Euclidean & 100.76  & $111.63 \pm 6.71s$ \\ 
	& Adam Euclidean PLU & 112.39   & $107.30 \pm  13.70s$ \\
	& Adam Manifold & 118.40  & $110.35 \pm  7.19s$ \\
	& ACClip Manifold (proposed) & 106.90 & \boldmath{$106.44 \pm 9.02s$} \\
	& CD-FNMRS & \textbf{97.08}  & $254.04 \pm 120.49s$ \\   
    \hline
    \multicolumn{4}{@{}l}{$^{\#}$ \scriptsize{Manifold ADAM has an unacceptable convergence, hence is not marked.}}
  \end{tabular}
\end{table}

\subsection{Synthetic Data}\label{simdata}
To compare the efficacy of different approaches in recovering genuine underlying distribution parameters, we draw samples from GMMs with known means and covariance matrices. As the separation of the components plays a crucial role in the learning procedures, data is generated using random sampling from randomly generated GMMs. The separation is controlled by this inequality $\forall_{i \neq j}\, \|\mu_i - \mu_j\|\, \geq \,c\,\underset{i,j}{\text{max}}\{\mathnormal{tr}(\Sigma_i), \mathnormal{tr}(\Sigma_i)\}$, where the separation control variable \textbf{\(c\)} is set to three thresholds 0.1, 1, and 5, for low, medium, and high separations. 
All of the synthesized datasets have five components (\(K\)) and are generated in five-dimensional space (\(n\)). Another aspect of performance can be presented in how many data points are needed to achieve convergence; to this end, datasets have been produced in three different sizes; \(10n^2\), \(100n^2\), and \(1000n^2\). In addition, The covariance matrices have an eccentricity of \(e = 10\), defined as the ratio of the largest to the smallest eigenvalue, and the batch size for stochastic methods is set to be fixed at 16.

There are three main error indices for evaluation, the averaged negative log-likelihood over ten different datasets -in Tables \ref{tab:sim_rand_data} and \ref{tab:sim_orth_data} named \textit{Avg. NLL}, the summation of frobenius norms of differences between the estimated covariance matrices and nominal covariance matrices over all components -named \textit{Cov. Err}, and the summation of cosine similarity distance of differences between the estimated mean vectors and nominal mean vectors over all components -named \textit{Mean Err}.

\subsubsection{Random Data}\label{randdata}
In this experiment, data are sampled from GMMs, for which there are no special constraints on the shape of the covariance matrices. The detailed results can be found in Table \ref{tab:sim_rand_data}; in general, online methods compete with the batch method. The proposed method and the ACClip SGD were shown to be the best online methods in high and mid separation respectively and they perform almost similarly. With large and medium dataset sizes in high and low separation categories, the NLL of online methods shows improvement upon the EM results. Moreover, the results for high and mid separation in small dataset sizes favor the EM-based algorithm. Given that we have 5 clusters and 250 samples in the entire dataset, it is predictable that using all samples at once may yield a better likelihood compared to using an online method. Also, the new manifold optimization method outperforms the ADAM version almost in all cases.

The convergence behavior of covariance errors in Table \ref{tab:sim_rand_data} shows that in case of acceptable convergence, smaller errors can be obtained by the online methods. On the other hand, the mean vector errors are lower for the online methods in the cases of high separation or a large number of data points.

\subsubsection{Orthogonal Data}\label{orthdata}
In this part, we conduct an experiment similar to the one in the previous section with the difference in the shape of the covariance matrix. The covariance matrices have the same structure as \(UD_kU^{\mathsf{T}}\) with orthogonal \(U\); hence the eccentricity factor can be defined on the diagonal elements of the matrix \(D_k\). The detailed results are presented in Table \ref{tab:sim_orth_data}.

In all the synthetic orthogonal cases, the online methods require fewer epochs for convergence. Also, they achieve a competitive or better convergence of NLL except in the circumstance of the high separation with a small dataset. The EM-based algorithm appears to be dramatically slow in the first part, which uses coordinate descent. 

The reported covariance errors in Table \ref{tab:sim_orth_data} highlight the excellence of the new manifold optimization algorithm outperforming all other methods in terms of this error, with the exception of small dataset sizes. The decrease in the separation mode impacts the performance of all online methods.

\begin{figure*}[h!]
  \centering
  \begin{subfigure}{0.32\linewidth}
    \includegraphics[width=1 \linewidth]{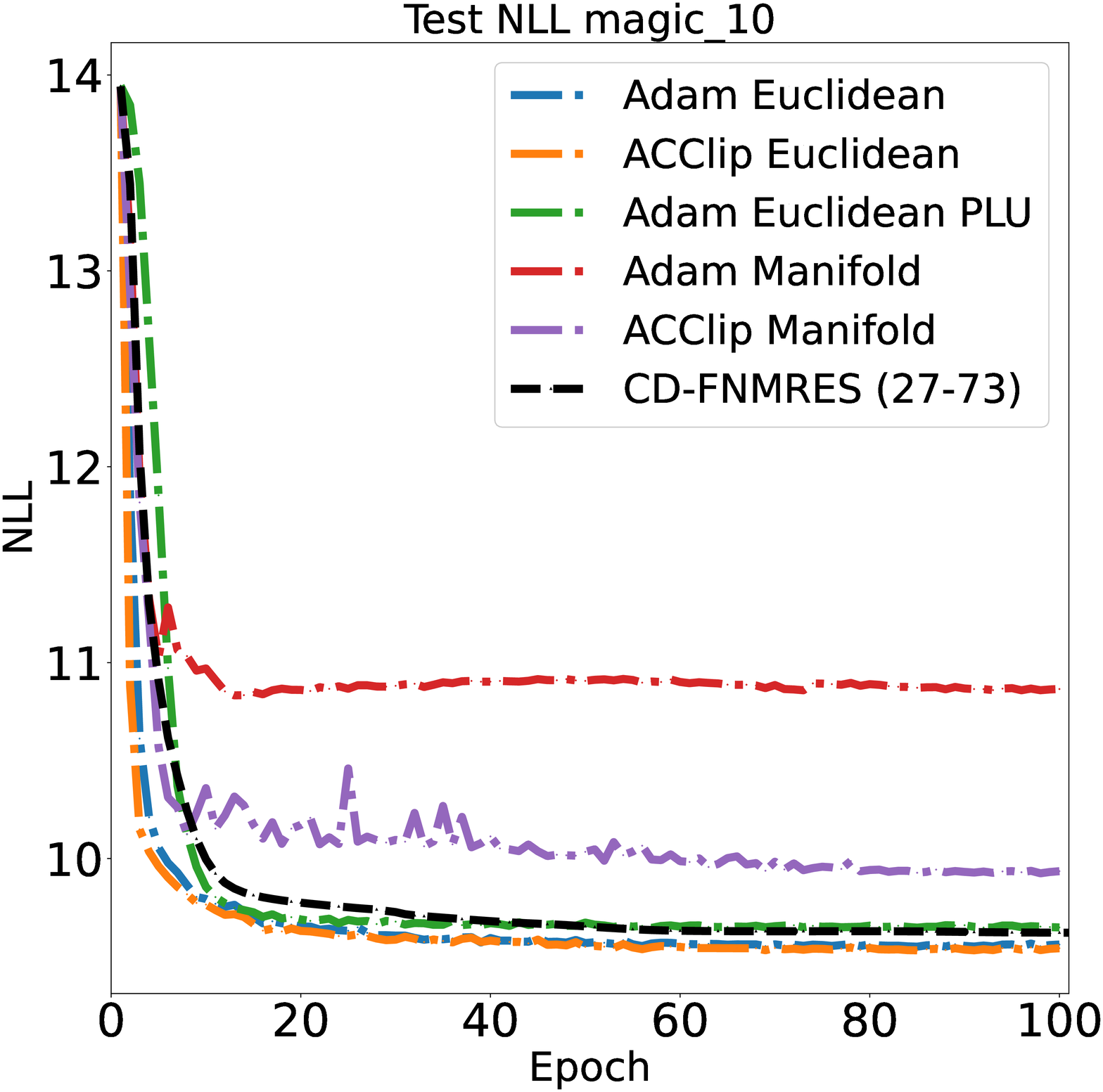}
    \caption{MAGIC dimension of 10}
    \label{fig:magic}
  \end{subfigure}
  \hfill
  \begin{subfigure}{0.32\linewidth}
    \includegraphics[width=1\linewidth]{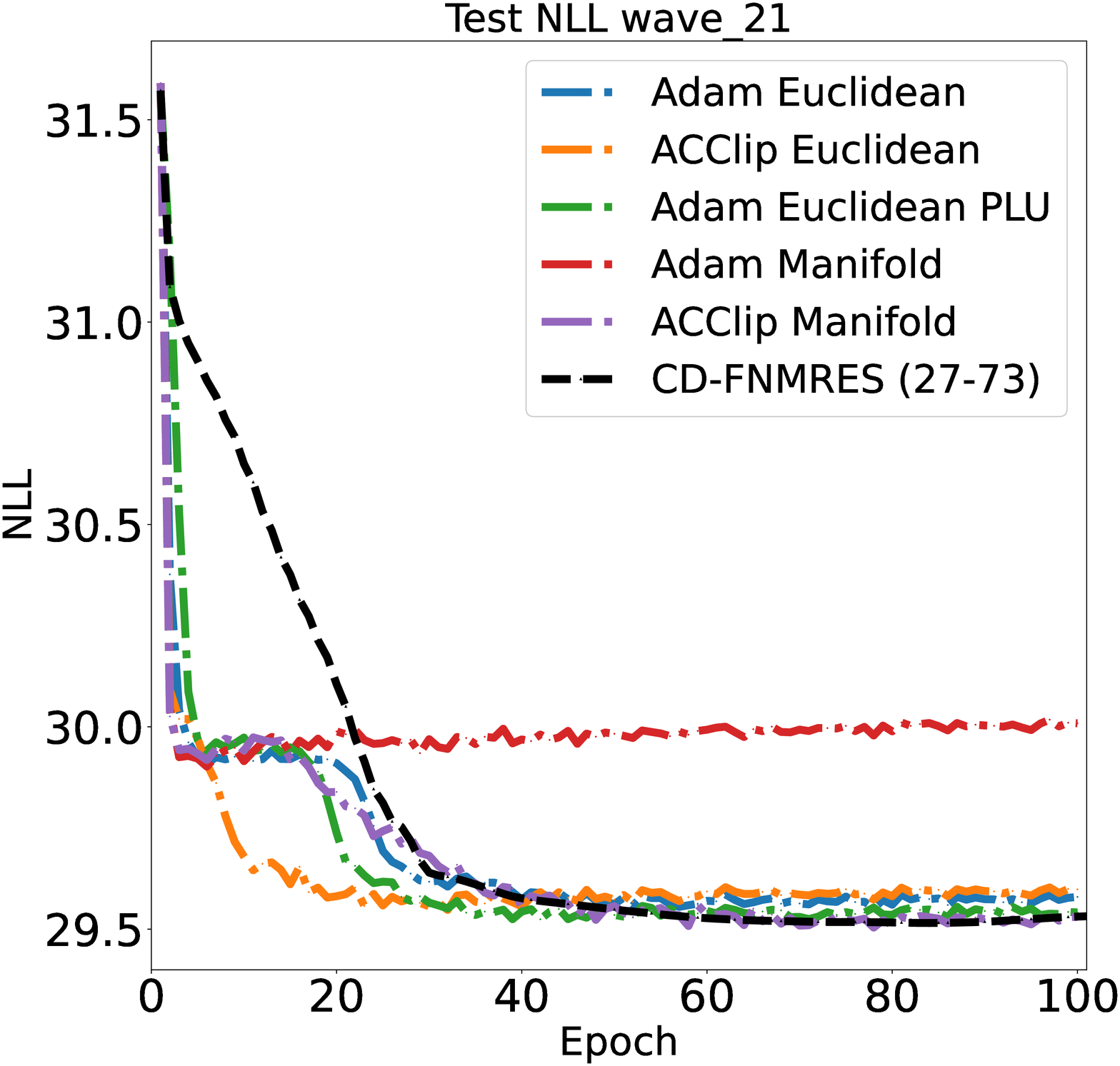}
    \caption{WAVE dimension of 21}
    \label{fig:wave}
  \end{subfigure}
  \begin{subfigure}{0.32\linewidth}
    \includegraphics[width=1\linewidth]{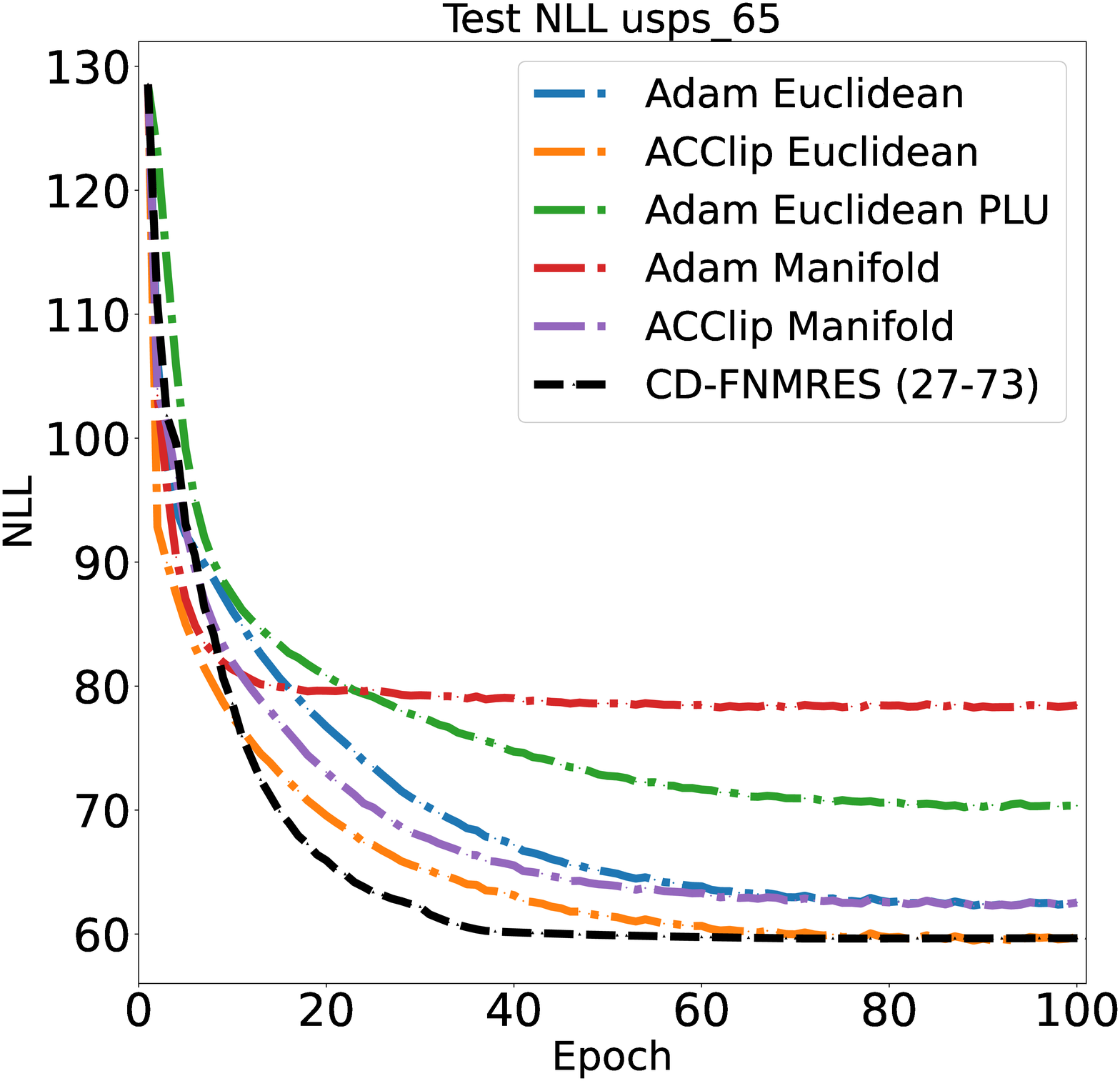}
    \caption{USPS dimension of 65}
    \label{fig:usps}
  \end{subfigure}
  \begin{subfigure}{0.32\linewidth}
    \includegraphics[width=1\linewidth]{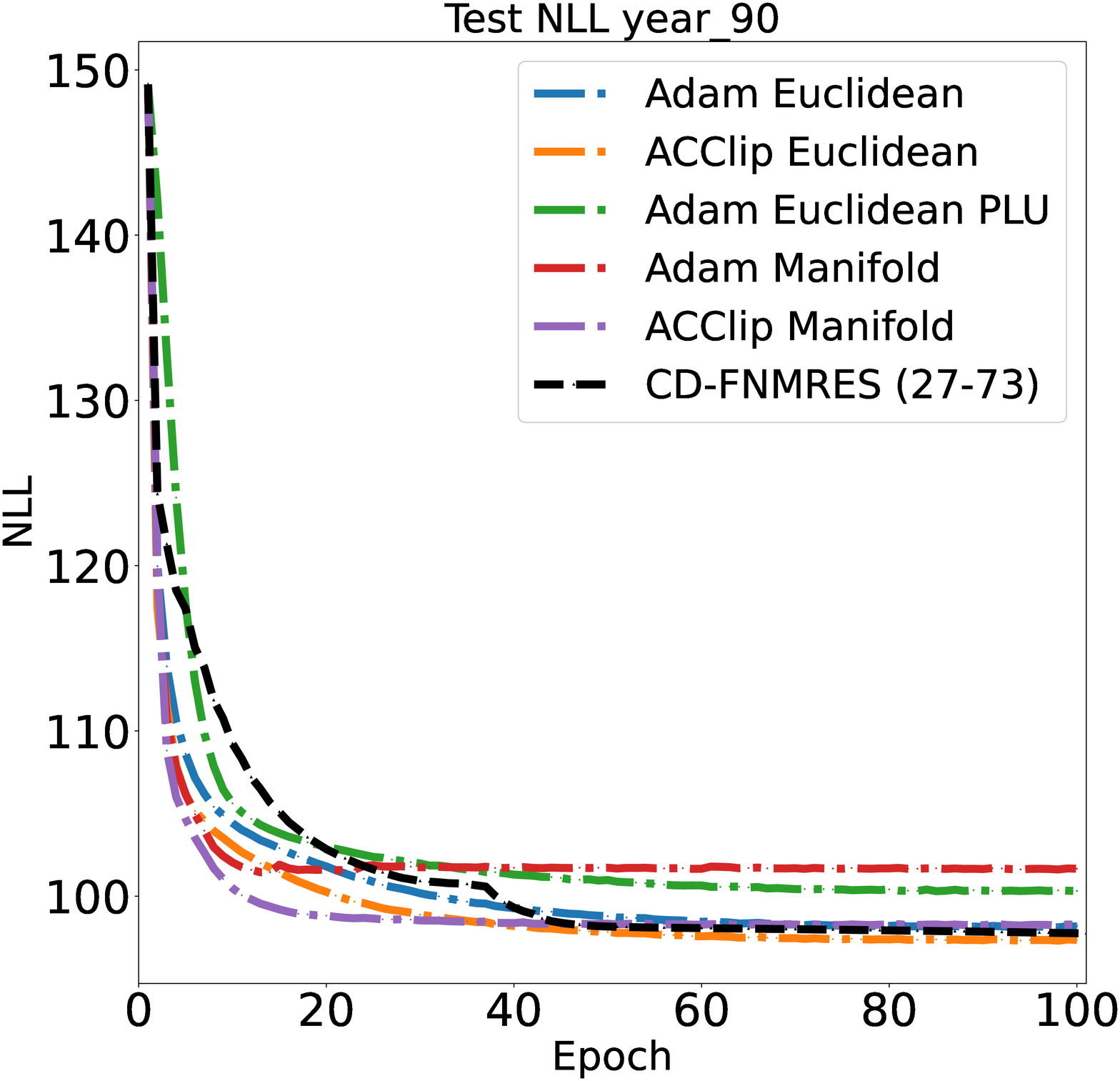}
    \caption{YEAR dimension of 90}
    \label{fig:year}
  \end{subfigure}
  \hfill
  \begin{subfigure}{0.32\linewidth}
    \includegraphics[width=1\linewidth]{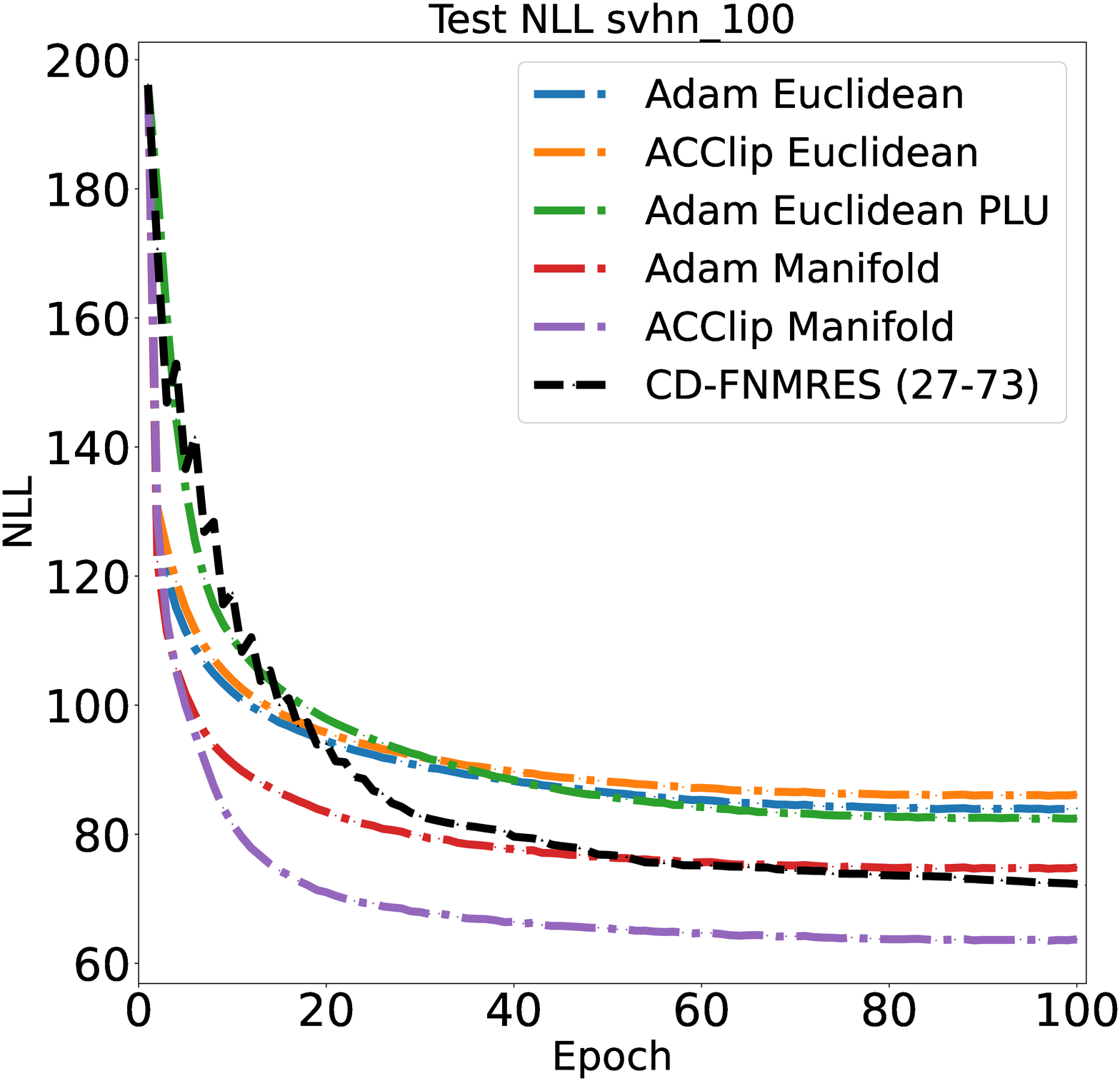}
    \caption{SVHN dimension of 100}
    \label{fig:svhn}
  \end{subfigure}
  \begin{subfigure}{0.32\linewidth}
    \includegraphics[width=1\linewidth]{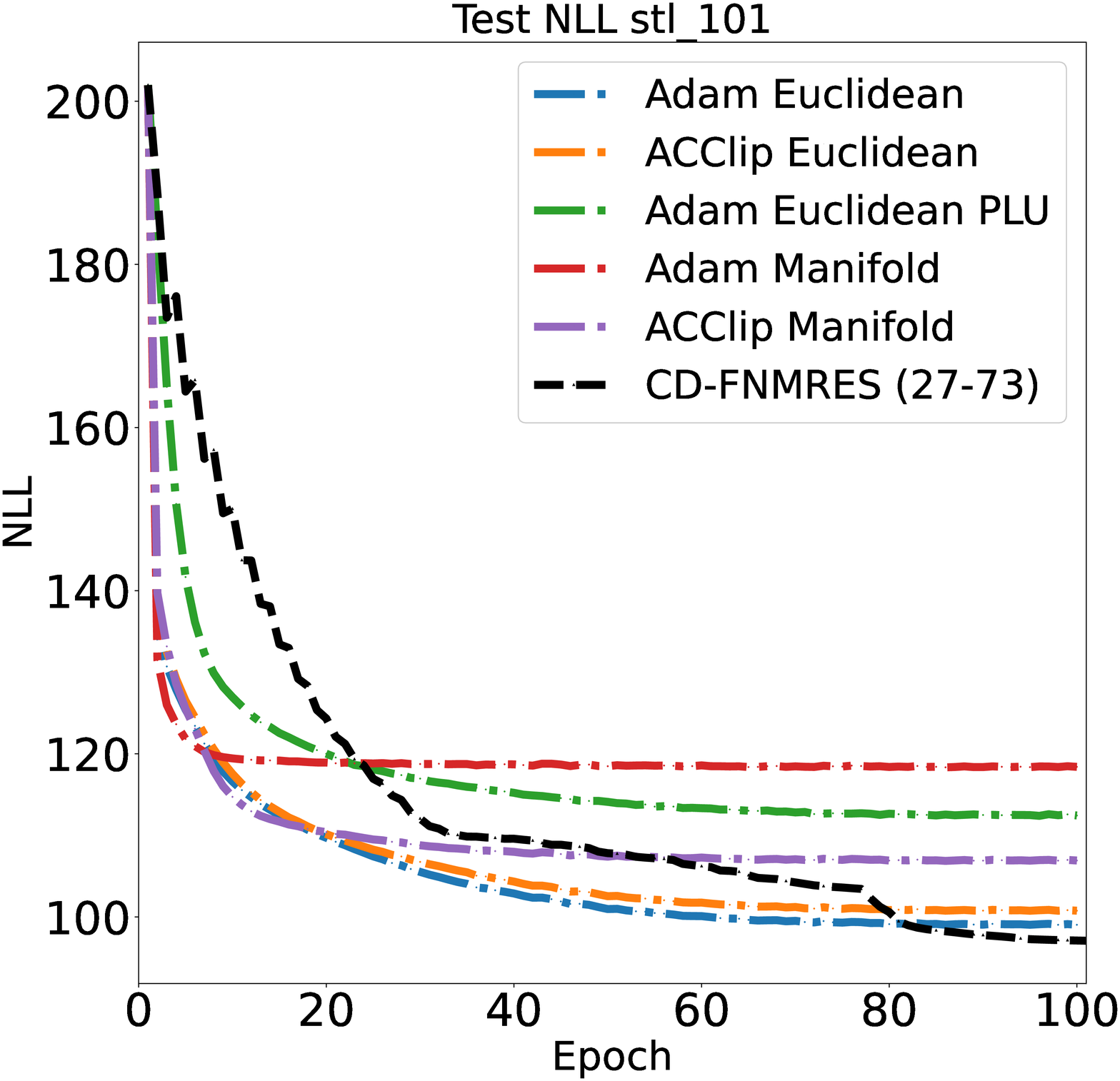}
    \caption{STL-10 dimension of 101}
    \label{fig:stl}
  \end{subfigure}  
  \caption{The Negative Log-Likelihood for the test partition of real datasets. The first four algorithms are the gradient-based methods in both Euclidean and manifold domains, and CD-FNMRES (27-73) indicates Coordinate Descent-Fast Newton Minimum ReSidual EM algorithm (the EM baseline) which takes 27 steps of coordinate descent and 73 steps of fast Newton minimum residual. \textbf{Note}: ``ACClip Manifold'' indicates the proposed Riemannian algorithm.}
  \label{fig:nll real data}
\end{figure*}
\subsection{Real Data}\label{realdatasec}
For the last set of experiments, we conduct a fair procedure (in the sense of computation power allocation and initialization) to evaluate the online methods (including ours) and compare them with the EM-based method over real data with a large number of components. 

Since the GMM has been known not to be desirable with image data and computer vision tasks, we include image datasets for our experiments.
The datasets used in this section are:
``Magic Gamma Telescope''\cite{bock2004methods},
 ``Wave Energy Converters''\cite{mann2007ceto},
 ``SVHN'' \cite{netzer2011reading}, ``STL-10'' \cite{coates2011analysis}, ``USPS'' \cite{hull1994database},
 and ``YearPredictionMSD'' available via \url{https://archive.ics.uci.edu/ml/datasets}.
To lift the structural variable (hyperparameter of GMMs) obligation, the number of Gaussian components in each example is the same as the dimension of the dataset itself. 

The results of Table \ref{tab:peformance real data} and Figure \ref{fig:nll real data} portray this fact, at least one of the online algorithms fulfills a better convergence than the EM-based algorithm. Except for the STL-10 dataset, in which the EM-based method is inferior to the online method before the last 20 epochs. Also, in all cases, the proposed ACCliping SGD on the manifold is superior to the ADAM version and outperforms others on the SVHN dataset. Furthermore, in all examples, the best online method converges to an optimal point with fewer epochs, excluding the USPS dataset. Moreover, The time consumption per epoch/iteration becomes in favor of online methods by increasing the dimension (up to four times faster).

\section{Conclusion}
\label{others}

In this paper, we proposed a framework for online GMM learning by using the first-order stochastic methods in high-dimensional spaces with large number of components. The flexibly-tied factorization of covariance matrices (\(\Sigma^{-1}_k = UD_kU^{\mathsf{T}}\)) was used to achieve this goal. We used the orthogonality constraint of the matrix \(U\) in this parameter-sharing scheme. Our main contribution was proposing a new stochastic manifold optimizer (clipping SGD on the product manifold of \(\mathcal{SO}(n)\) and several Euclidean spaces) to handle the orthogonality constraint on the product manifold of GMM parameters. The empirical experiments verified that the proposed online routine converged faster (in the sense of the number of epochs), with less computation time in higher dimensions, outperforming the EM-based method in various scenarios. The simplicity of implementation makes this framework practically suitable in domains that need an online estimation of large-scale GMMs.

Our results contradicted the primary assumption that orthogonality may restrict the model performance. By achieving even better time performances than its Euclidean counterpart, our optimizer can be used in more complex models such as deep neural networks, preserving orthogonality during the training. At last, in Euclidean-based scenarios, the utilization of PLU factorization for matrix \(U\) to prune the computation of covariance matrix determinants led to improved time performance but its convergence was inferior to some other algorithms.

\bibliographystyle{unsrt} 
\bibliography{references}

\begin{thebibliography}{10}

\bibitem{mclachlan2019finite}
Geoffrey~J McLachlan, Sharon~X Lee, and Suren~I Rathnayake.
\newblock Finite mixture models.
\newblock {\em Annual Review of Statistics and its Application}, 6:355--378, 2019.

\bibitem{khansari2011learning}
Seyed~Mohammad Khansari-Zadeh and Aude Billard.
\newblock Learning stable nonlinear dynamical systems with {G}aussian mixture models.
\newblock {\em IEEE Transactions on Robotics}, 27(5):943--957, 2011.

\bibitem{kalayeh2019training}
Mahdi~M Kalayeh and Mubarak Shah.
\newblock Training faster by separating modes of variation in batch-normalized models.
\newblock {\em IEEE Transactions on Pattern Analysis and Machine Intelligence}, 42(6):1483--1500, 2019.

\bibitem{kolouri2018sliced}
Soheil Kolouri, Gustavo~K Rohde, and Heiko Hoffmann.
\newblock Sliced {W}asserstein distance for learning {G}aussian mixture models.
\newblock In {\em IEEE/CVF Conference on Computer Vision and Pattern Recognition}, pages 3427--3436, 2018.

\bibitem{richardson2018gans}
Eitan Richardson and Yair Weiss.
\newblock On {GAN}s and {GMM}s.
\newblock In {\em Advances in Neural Information Processing Systems}, volume~31, 2018.

\bibitem{dempster1977maximum}
Arthur~P Dempster, Nan~M Laird, and Donald~B Rubin.
\newblock Maximum likelihood from incomplete data via the {EM} algorithm.
\newblock {\em Journal of the Royal Statistical Society: Series B (Methodological)}, 39(1):1--22, 1977.

\bibitem{redner1984mixture}
Richard~A Redner and Homer~F Walker.
\newblock Mixture densities, maximum likelihood and the {EM} algorithm.
\newblock {\em SIAM Review}, 26(2):195--239, 1984.

\bibitem{mclachlan2007algorithm}
Geoffrey~J McLachlan and Thriyambakam Krishnan.
\newblock {\em The {EM} algorithm and extensions}.
\newblock John Wiley \& Sons, 2007.

\bibitem{neal1998view}
Radford~M Neal and Geoffrey~E Hinton.
\newblock A view of the {EM} algorithm that justifies incremental, sparse, and other variants.
\newblock In {\em Learning in Graphical Models}, pages 355--368. Springer, 1998.

\bibitem{huda2009stochastic}
Shamsul Huda, John Yearwood, and Roberto Togneri.
\newblock A stochastic version of expectation maximization algorithm for better estimation of hidden {Markov} model.
\newblock {\em Pattern Recognition Letters}, 30(14):1301--1309, 2009.

\bibitem{asheri2021new}
Hadi Asheri, Reshad Hosseini, and Babak~Nadjar Araabi.
\newblock A new {EM} algorithm for flexibly tied {GMM}s with large number of components.
\newblock {\em Pattern Recognition}, 114:107836, 2021.

\bibitem{jin2016local}
Chi Jin, Yuchen Zhang, Sivaraman Balakrishnan, Martin~J Wainwright, and Michael~I Jordan.
\newblock Local maxima in the likelihood of {G}aussian mixture models: Structural results and algorithmic consequences.
\newblock In {\em Advances in Neural Information Processing Systems}, volume~29, 2016.

\bibitem{vanderbei2000formulating}
Robert~J Vanderbei and Hande~Yurttan Benson.
\newblock On formulating semidefinite programming problems as smooth convex nonlinear optimization problems.
\newblock Technical report, Center for Discrete Mathematics \& Theoretical Computer Science, 2000.

\bibitem{rahmani2020estimation}
Donya Rahmani, Mahesan Niranjan, Damien Fay, Akiko Takeda, and Jacek Brodzki.
\newblock Estimation of {Gaussian} mixture models via tensor moments with application to online learning.
\newblock {\em Pattern Recognition Letters}, 131:285--292, 2020.

\bibitem{hosseini2015matrix}
Reshad Hosseini and Suvrit Sra.
\newblock Matrix manifold optimization for {G}aussian mixtures.
\newblock In {\em Advances in Neural Information Processing Systems}, volume~28, pages 910--918, 2015.

\bibitem{hosseini2020alternative}
Reshad Hosseini and Suvrit Sra.
\newblock An alternative to {EM} for {G}aussian mixture models: batch and stochastic {R}iemannian optimization.
\newblock {\em Mathematical Programming}, 181(1):187--223, 2020.

\bibitem{amari1998natural}
Shun-Ichi Amari.
\newblock Natural gradient works efficiently in learning.
\newblock {\em Neural Computation}, 10(2):251--276, 1998.

\bibitem{bouguila2005using}
Nizar Bouguila and Djemel Ziou.
\newblock Using unsupervised learning of a finite {Dirichlet} mixture model to improve pattern recognition applications.
\newblock {\em Pattern Recognition Letters}, 26(12):1916--1925, 2005.

\bibitem{gales1999semi}
Mark~JF Gales.
\newblock Semi-tied covariance matrices for hidden {Markov} models.
\newblock {\em IEEE Transactions on Speech and Audio Processing}, 7(3):272--281, 1999.

\bibitem{biernacki2006model}
Christophe Biernacki, Gilles Celeux, G{\'e}rard Govaert, and Florent Langrognet.
\newblock Model-based cluster and discriminant analysis with the {MIXMOD} software.
\newblock {\em Computational Statistics \& Data Analysis}, 51(2):587--600, 2006.

\bibitem{kingma2018glow}
Durk~P Kingma and Prafulla Dhariwal.
\newblock Glow: Generative flow with invertible 1x1 convolutions.
\newblock In {\em Advances in Neural Information Processing Systems}, volume~31, 2018.

\bibitem{salakhutdinov2003optimization}
Ruslan Salakhutdinov, Sam~T Roweis, and Zoubin Ghahramani.
\newblock Optimization with {EM} and expectation-conjugate-gradient.
\newblock In {\em the 20th International Conference on Machine Learning}, pages 672--679, 2003.

\bibitem{jordan1994hierarchical}
Michael~I Jordan and Robert~A Jacobs.
\newblock Hierarchical mixtures of experts and the {EM} algorithm.
\newblock {\em Neural Computation}, 6(2):181--214, 1994.

\bibitem{kingma2014adam}
Diederik~P Kingma and Jimmy Ba.
\newblock Adam: A method for stochastic optimization.
\newblock In {\em International Conference on Learning Representations}, 2014.

\bibitem{zhang2020adaptive}
Jingzhao Zhang, Sai~Praneeth Karimireddy, Andreas Veit, Seungyeon Kim, Sashank Reddi, Sanjiv Kumar, and Suvrit Sra.
\newblock Why are adaptive methods good for attention models?
\newblock In {\em Advances in Neural Information Processing Systems}, volume~33, pages 15383--15393, 2020.

\bibitem{celeux1995gaussian}
Gilles Celeux and G{\'e}rard Govaert.
\newblock Gaussian parsimonious clustering models.
\newblock {\em Pattern Recognition}, 28(5):781--793, 1995.

\bibitem{absil2009optimization}
Pierre-Antoine Absil, Robert Mahony, and Rodolphe Sepulchre.
\newblock {\em Optimization algorithms on matrix manifolds}.
\newblock Princeton University Press, 2009.

\bibitem{li2020efficient}
Jun Li, Li~Fuxin, and Sinisa Todorovic.
\newblock Efficient {R}iemannian optimization on the {S}tiefel manifold via the {C}ayley transform.
\newblock In {\em International Conference on Learning Representations}, 2020.

\bibitem{bonnabel2013stochastic}
Silvere Bonnabel.
\newblock Stochastic gradient descent on {R}iemannian manifolds.
\newblock {\em IEEE Transactions on Automatic Control}, 58(9):2217--2229, 2013.

\bibitem{bock2004methods}
Rudolf~K Bock, Markus Chilingarian, Ashot Agassiand~Gaug, Frantisek Hakl, Thomas Hengstebeck, Marcel Ji{\v{r}}ina, Jan Klaschka, Emil Kotr{\v{c}}, Petr Savick{\`y}, Sherry Towers, Anthony Vaiciulis, and Wittek Wolfgang.
\newblock Methods for multidimensional event classification: a case study using images from a {C}herenkov {G}amma-ray telescope.
\newblock {\em Nuclear Instruments and Methods in Physics Research Section A: Accelerators, Spectrometers, Detectors and Associated Equipment}, 516(2-3):511--528, 2004.

\bibitem{mann2007ceto}
Laurence~D Mann, Alan~R Burns, and Michael~E Ottaviano.
\newblock {CETO}, a carbon free wave power energy provider of the future.
\newblock In {\em the 7th European Wave and Tidal Energy Conference}, 2007.

\bibitem{netzer2011reading}
Yuval Netzer, Tao Wang, Adam Coates, Alessandro Bissacco, Bo~Wu, and Andrew~Y Ng.
\newblock Reading digits in natural images with unsupervised feature learning.
\newblock In {\em Advances in Neural Information Processing Systems Workshop on Deep Learning and Unsupervised Feature Learning}, 2011.

\bibitem{coates2011analysis}
Adam Coates, Andrew Ng, and Honglak Lee.
\newblock An analysis of single-layer networks in unsupervised feature learning.
\newblock In {\em the 14th International Conference on Artificial Intelligence and Statistics}, volume~15, pages 215--223, 2011.

\bibitem{hull1994database}
Jonathan~J. Hull.
\newblock A database for handwritten text recognition research.
\newblock {\em IEEE Transactions on Pattern Analysis and Machine Intelligence}, 16(5):550--554, 1994.

\end{thebibliography}

\end{document}